%% file: acl2023.tex
\pdfoutput=1

\documentclass[11pt]{article}

\usepackage{ACL2023}
\usepackage{subfiles}
\usepackage{times}
\usepackage{latexsym}
\usepackage{graphicx}
\usepackage{booktabs}
\usepackage{multirow}
\usepackage{amsmath}
\usepackage{tabularx}

\usepackage{tcolorbox}
\usepackage{multicol}

\newtcolorbox{promptbox}{colback=blue!5!white, colframe=blue!75!black, boxrule=0.5pt, arc=3mm, width=\columnwidth}

\newtcolorbox{promptbox2}{colback=red!5!white, colframe=blue!75!black, boxrule=0.5pt, arc=3mm, width=\columnwidth}

\newtcolorbox{multicolpromptbox}[1][]{float*=t, colback=yellow!10, colframe=yellow!50!black, boxrule=0.5pt, arc=3mm, width=\textwidth, #1}

\usepackage{array}
\usepackage[T1]{fontenc}

\usepackage[utf8]{inputenc}

\usepackage{microtype}

\usepackage{inconsolata}

\title{ Follow-up Question Generation\\ For Enhanced Patient-Provider Conversations}

\author{Joseph Gatto, Parker Seegmiller, Timothy Burdick, Inas S. Khayal, Sarah DeLozier, Sarah M. Preum\\Department of Computer Science, Dartmouth College\\
\texttt{\{joseph.m.gatto.gr, sarah.masud.preum\}@dartmouth.edu}}

\begin{document}
\maketitle 
\begin{abstract}
Follow-up question generation is an essential feature of dialogue systems as it can reduce conversational ambiguity and enhance modeling complex interactions. Conversational contexts often pose core NLP challenges such as (i) extracting relevant information buried in fragmented data sources, and (ii) modeling parallel thought processes. These two challenges occur frequently in medical dialogue as a doctor asks questions based not only on patient utterances but also their prior EHR data and current diagnostic hypotheses. Asking medical questions in \textit{asynchronous conversations} compounds these issues as doctors can only rely on static EHR information to motivate follow-up questions.

To address these challenges, we introduce \textbf{FollowupQ}, a novel framework for enhancing asynchronous medical conversation.
FollowupQ is a multi-agent framework that processes patient messages and EHR data to generate personalized follow-up questions, clarifying patient-reported medical conditions. FollowupQ reduces requisite provider follow-up communications by 34\%. It also improves performance by 17\% and 5\% on real and synthetic data, respectively. We also release the first public dataset of asynchronous medical messages with linked EHR data alongside 2,300 follow-up questions written by clinical experts for the wider NLP research community.

\end{abstract}

\section{Introduction}

\subfile{sections/1_Introduction}

\section{Related Work}
\subfile{sections/2_RelatedWork}

\subfile{sections/4_Methods_Parker_Rewrite}

\subfile{sections/5_ExperimentalSetup}

\section{Results}\label{sec:results}

\subfile{sections/6_Results}

\subfile{sections/7_Discussion}

\bibliography{custom}

\appendix 
\subfile{sections/Appendix}

\end{document}

%% file: sections/1_Introduction.tex
\begin{figure}[t]
    \centering
    \includegraphics[width=\columnwidth]{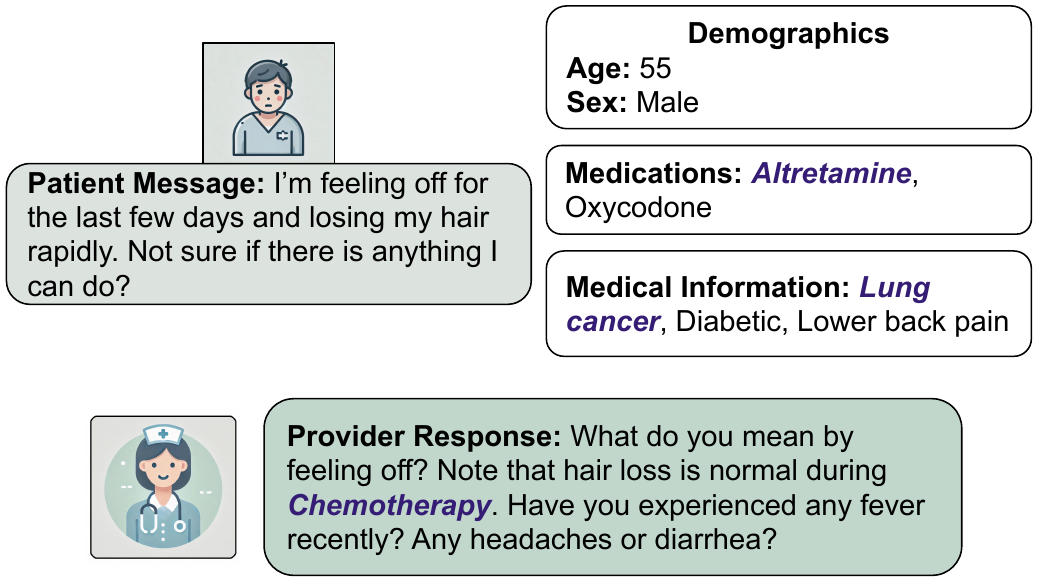}
    \caption{We demonstrate the core challenges of asynchronous follow-up question generation. Providers need to consider complex fragmented data sources to generate \textit{multiple} follow-up questions reflecting parallel thought processes.}
    \label{fig:FIG1_Example}
\end{figure}

Asking relevant, useful follow-up questions while conversing fosters deeper understanding, and ensures meaningful and productive conversations. Thus neural dialogue systems may benefit from the ability to generate good follow-up questions, eliciting required information and reducing conversational ambiguity in real-time \cite{yi2024surveyrecentadvancesllmbased}. Follow-up Question Generation has been studied in domains such as social media \cite{meng2023followupqg, liu-etal-2025-superficial}, healthcare \cite{mediQ}, document understanding \cite{ko-etal-2020-inquisitive}, and conversational surveys \cite{ge-etal-2023-ask}. 

Generating relevant, meaningful follow-up questions is a non-trivial NLP task as it often involves gathering relevant information that is fragmented across multiple sources \cite{chen-etal-2021-action} or requires complex parallel thought processes \cite{lee-etal-2024-towards, cao2024diaggptllmbasedmultiagentdialogue}. 
For example, in patient-provider communication, providers need to (i) consider patient utterances while attending to information scattered throughout the patient's Electronic Health Record (EHR) \cite{tu2024conversationaldiagnosticai} and (ii) consider numerous different thought processes while formulating a patient's diagnosis \cite{mcduff2023accuratedifferentialdiagnosislarge, huynh2023understanding}. A similar challenge arises in customer service interactions, where support agents must (i) integrate information from a customer’s message while referencing their past interactions, purchase history, or account details \cite{Xu_2024, MISISCHIA2022421} and (ii) simultaneously consider multiple resolution strategies, such as troubleshooting steps, rerouting to other support, or refund policies \cite{pi2024contactcomplexitycustomerservice}.

Prior works have studied the information-seeking ability of dialogue systems including LLM-based systems in the context of \textit{synchronous} medical conversation --- where provider and patient engage in real-time multi-turn conversation \cite{mediQ, healthQ, stanfordHPIQ}. However, to the best of our knowledge, there has been limited exploration of LLMs as information-seeking agents in the context of \textit{asynchronous} medical dialogue --- where two parties exchange online messages at their convenience. Asynchronous medical conversations are getting increasingly popular with telehealth services \cite{shaver2022state} and increased use of online patient portals \cite{hansen2023impact} and have unique traits which demand targeted NLP solutions.

First, when one messages their doctor asynchronously, it is common for them to assume the reader of their message has significant knowledge of their personal and medical background \cite{gatto2024incontextlearningpreservingpatient}. This data characteristic requires NLP systems to rely heavily on fragmented data in the EHR to fill-in missing context --- as there is no live patient for information solicitation. Second, when a provider solicits additional information from a patient in this setting, they need to generate a \textit{\textbf{list}} of follow-up questions to rule out a broad range of possible diagnoses all at once to reduce the need for additional follow-up communication (e.g. calling the patient). This contrasts the synchronous communication paradigm, where patients are usually asked one question at a time.

An NLP system capable of automatically generating relevant follow-up questions when patients report their medical symptoms has significant real-world benefits. It can reduce the volume of follow-up communications required from providers, thereby alleviating a key contributor to provider burnout — the burden of asynchronous messaging \cite{death_by_portal, budd2023burnout}. However, this task remains largely unexplored in the NLP community due to the absence of publicly available datasets and effective evaluation methods.

In this study, we solve these two problems by introducing and formalizing a new task, \textbf{Follow-up Question Generation} for Asynchronous Patient-Provider Conversations. We release \textbf{FollowupBench}, the first public dataset of 250 semi-synthetic patient data samples, containing patient messages and EHR data. This dataset contains over \textbf{2,300 follow-up questions} composed by a team of 9 primary care providers (including doctors, nurses, and medical residents) at a regional university medical center. Our study additionally explores follow-up question generation on a set of 150 real patient messages with ground truth questions extracted from responses written by real providers during their normal workflow. We introduce a novel evaluation metric, Requested Information Match, which formalizes the task of comparing predicted vs ground-truth question sets in the context of asynchronous medical conversations using LLM-as-Judge \cite{llm_as_judge}. 

However, our experiments demonstrate that \textit{off-the-shelf LLMs struggle to generate follow-up questions} that align with ground-truth questions written by real providers. We address this limitation in this setting by introducing \textbf{FollowupQ}, a multi-agent framework for personalized follow-up question generation in patient messages. FollowupQ utilizes an agentic divide-and-conquer strategy to alleviate the complexity of mapping multiple complex data sources to a set of follow-up questions. FollowupQ agents, developed in collaboration with a team of clinical experts, collaborate to emulate complex clinical thought processes spanning a broad range of clinical inquiry. The contributions of this work can be summarized as follows:

\begin{enumerate}
    \item We introduce FollowupQ: A novel multi-agent framework for follow-up question generation for patient message enhancement. FollowupQ outperforms baselines by 17\% and 5\% on real and synthetic datasets, respectively.
   
    \item We release the first asynchronous medical messaging dataset with both patient messages and EHR data --- mimicking a real-world environment. Our open-source dataset contains 2,300 questions from real providers covering a wide variety of medical conditions. 
    
    \item We introduce a novel evaluation metric, Requested Information Match (RIM), which uses LLM-as-Judge to help quantify the potential reduction in provider workload. 
\end{enumerate}

%% file: sections/2_RelatedWork.tex
\subsection{Follow-Up Question Generation}

The importance of generating high-quality follow-up question generations has been studied throughout conversational NLP. For example, there have been a variety of studies on information-seeking systems for social medial conversations \cite{meng2023followupqg, liu-etal-2025-superficial}, and conversational surveys \cite{ge-etal-2023-ask}. However, unlike FollowupQ these works focus on generating one question at a time. Asking follow-up questions in medical conversations has been explored in the related context of synchronous dialogue generation. For example, \citet{stanfordHPIQ} explore the ability of LLMs to ask patients questions towards eliciting pertinent background details about their visit. \citet{healthQ} transform medical dialogue datasets into a format suitable for medical question generation evaluation. \citet{mediQ} transform Medical-QA datasets such as Med-QA \cite{jin2021disease} to explore the capacity of LLMs to ask follow-up questions that lead to a medical diagnosis. 

None of these studies explore asynchronous conversations such as those which occur in online patient portals and telehealth services. Additionally, our work instead focuses on generation of \textit{sets} of questions, enabling comparison to ground-truth questions from real portal message interactions. To the best of our knowledge, the only other work to explore follow-up question generation in a similar setting is \citet{vander} --- but at an extremely small scale (n=7 synthetic samples) and without any utilization of linked EHR data.

\subsection{Multi-Agent Systems in Healthcare}
Recently LLM-based Multi-Agent systems have been shown to provide significant performance increases across a broad range of tasks and domains including healthcare \cite{guo2024largelanguagemodelbased} . Frameworks such as MedAgents \cite{tang-etal-2024-medagents}, RareAgents \cite{chen2024rareagentsautonomousmultidisciplinaryteam}, MDAgents \cite{kim2024mdagentsadaptivecollaborationllms}, and TriageAgent \cite{lu-etal-2024-triageagent} -- all leverage collaborative multi-round discussion between multiple LLM agents to perform medical decision making. Our framework, FollowupQ, is inspired by prior works as we also employ multiple agents for follow-up question generation. However, these prior systems are designed for medical decision-making, not information seeking, making them non-trivial to apply to our dataset. We further highlight the unique evaluation challenges addressed by FollowupQ in Section \ref{sec3}.

%% file: sections/4_Methods_Parker_Rewrite.tex
\begin{figure*}[!h]
    \centering
    \includegraphics[width=0.95\linewidth]{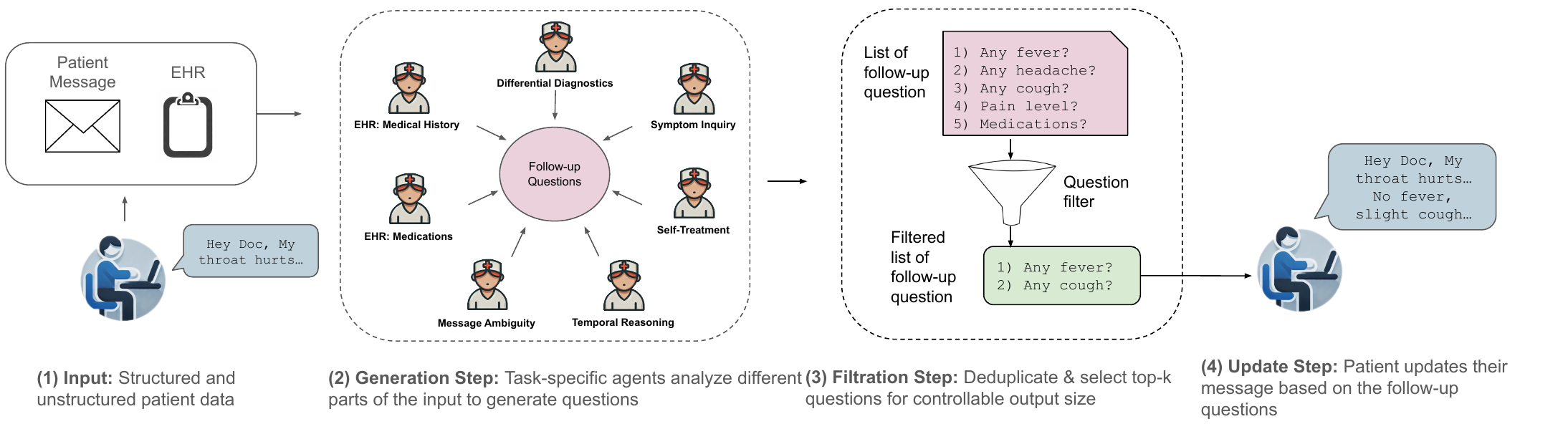}
    \caption{FollowupQ works by taking a patient message and a subset of their EHR and employing multiple LLM agents to explore diverse clinical thought processes --- producing a pool of follow-up questions from different perspectives. If desired, FollowupQ can then filter the output to a controllable question set size. }
    \label{fig:methods}
\end{figure*}

\section{Follow-Up Question Generation} \label{sec3}

\subsection{Problem Formalization}
\label{sec:task-formal} 

Consider a patient message $T$, their corresponding EHR $C$, and a text generator $f$ (e.g., an LLM or LLM-based framework).

$$f(T,C) = \hat{Q} = \{\hat{q_1}, \hat{q_2}, \dots, \hat{q_n}\}$$

The goal of the text generation system $f$ is to produce a set $\hat{Q}$ where each $\hat{q_i} \in \hat{Q}$ is a follow-up question to the patient's message. Crucially, $f$ exists in an asynchronous environment without real-time access to the patient and must generate all pertinent follow-up questions as a list.

We specifically define a patient's EHR record as $C = \{A, H, M\}$. This includes a patient's demographics $A$ (e.g. age and gender), medical history $H$ (e.g. problem list and recent medical encounters), and medication list $M$. Each component of $C$ is represented as a string in our framework. An example message with EHR data and provider response are shown in Appendix \ref{apn:example_data}.

\subsection{Evaluation Strategy}\label{sec:formal_metric}

In the context of the real-world use cases of asynchronous medical dialogue, the quality of the set of questions $\hat{Q} = \{\hat{q_1}, \dots, \hat{q_n}\}$ produced by $f$ is primarily determined by \textit{the reduction in dialogue turns required by the provider to make a diagnosis or recommendation.} To make their decision, providers must first have all the necessary information from the patient. To reduce provider workload, patient responses to questions in $\hat{Q}$ must contain at least the information requested in the ground truth question set $Q$. Thus, we design two metrics comparing generated questions $\hat{Q}$ against ground truth questions $Q$, to identify how well the information requested in $Q$ is covered by the information requested in $\hat{Q}$. 

Importantly, a generated question $\hat{q_i} \in \hat{Q}$ is considered as matching a ground truth question $q_j \in Q$ when they \textit{request the same information} (i.e., invoke similar responses),--- in addition to when they match exactly. For example, an LLM asking ``Do you have a cough or fever?''  will elicit a similar response to the doctor's question ``Have you been coughing?'' and thus should be considered a match. We employ LLM-as-judge \cite{llm_as_judge} to do this semantic matching. We present the validation of the LLM-as-judge process in Section \ref{sec:llm-as-judge}.

First, we introduce the \textbf{Requested Information Match (RIM)} metric.

\begin{equation}\label{eq:RIM}
    RIM(Q, \hat{Q}) = \frac{|Q \cap \hat{Q}|}{|Q|} 
\end{equation}

RIM measures the sample-wise percentage of provider-generated questions that are also generated by system $f$. RIM is a task-specific case of the Tversky Index \cite{tversky1977features}. RIM is notably distinct from other set comparison metrics, like Jaccard Similarity, as we do not penalize $f$ based on the size of the set of generated questions, $|\hat{Q}|$. This design choice is informed by our prior work with providers, as \textit{eliciting extra information helps providers}. For example, just because a provider only asked 3 questions to a patient, it does not mean there are no other useful inquiries to generate. Different providers might generate different list of questions based on their experience,  preferences, and current mental models, resulting in subjectivity in the ground truth. Thus requesting additional information helps a provider who may have forgotten to consider a certain outlying issue the patient may be facing and address the subjectivity of their thought process. 

Subsequently, we note that the first step towards solving asynchronous follow-up question generation is to optimize for coverage of ground truth questions (i.e. maximize RIM), with the secondary objective being controlling the size of the list of generated questions, $|\hat{Q}|$. In Section \ref{sec:filter}, we introduce a way to control $|\hat{Q}|$, i.e.the number of questions presented to a patient. However, as we discuss in Section \ref{sec:results}, RIM maximization is a difficult challenge for LLMs with unconstrained $|\hat{Q}|$. Thus our core metric is the average RIM score across all samples in our dataset. 

A RIM score of 1.0 denotes the case where system $f$ requested all information actually requested by the patient's real doctor. For each output with an RIM score of 1.0, we argue that $f$ has successfully reduced the number of clarification requests a provider needs to send by 1, assuming the patient responds to all questions. Conversely, while RIM scores below 1.0 are still suggestive of message improvement, they may not suggest any reduction in outgoing messages to be sent by a provider. 

Hence, we introduce a second metric \textbf{Message Reduction \% (MR\%)} which measures the percentage of samples where the RIM score = 1.0. Models with a higher MR\% have more real-world impact as they would lead to greater workload reductions. In the remainder of this section, we present the FollowupQ framework.

\section{FollowupQ Framework for Generating Follow-Up Questions}

\subsection{Design Motivation}\label{sec:design_motivation}
Our FollowupQ framework is summarized in Figure \ref{fig:methods}. Our methods were developed over a 12-month period in collaboration with a team of clinical experts who frequently communicate with patients in primary care. The experts include primary care physicians, medical residents, registered nurses, and triage nurses. Through a series of workshops and interviews with our team of experts, we discovered the following three thought processes essential to the asynchronous followup question generation process which we embed into FollowupQ. (i) EHR Reasoning: where providers obtain contextual knowledge from fragmented EHR data to guide their inquiry. (ii) Differential Diagnostics: Where providers develop a mental list of potential diagnoses or explanations that describe the patient's symptoms --- guiding their question formulation. (iii) Message Clarifications: Where providers ask a series of questions to fill in gaps in the patient's reported symptoms. 

\subsection{Followup Question Generation} 
Recall from Section \ref{sec:task-formal} that $T$ and $C$ denote a patient message and corresponding EHR record,     respectively. The objective of the generation step is to create a set of questions $f(T,C) = \hat{Q}$ where $\hat{Q} = \{\hat{q_1} \dots \hat{q_n} \}$. We build $\hat{Q}$ by taking the union of question sets generated by $m$ inquiry agents, i.e. $\hat{Q} = \hat{Q}_{{agent}_1} \dots \cup \dots \hat{Q}_{{agent}_m}$. This strategy allows us to use multiple agents to generate question sets corresponding to the parallel clinical thought processes outlined in Section \ref{sec:design_motivation}. $\hat{Q}$ is thus constructed using three high-level agents as described below: EHR reasoning agents, differential diagnostic agents, and message clarification agents.

\noindent \textbf{EHR Reasoning Agents: } 
EHR data is complex as the information related to a patient's current inquiry can be implicit or fragmented across different data tables, and fields. FollowupQ uses two EHR-specific reasoning agents to mitigate the challenges of generating questions from fragmented data sources. Specifically, for a given EHR record $C = \{A, H, M \}$ (see Section \ref{sec:task-formal} for definition of EHR elements), we define a medical history reasoning agent and a medication list reasoning agent. Each agent first extracts relevant pieces of EHR information concerning the patient's current inquiry $T$. This is critical as a patient's medical history and medication list can contain a lot of information that is not relevant to their current message.

{
\small
\begin{equation}
\begin{aligned}
    &I_{hist} = f(A, H, P_{extract_H}, T)\\
    &I_{med} = f(M, P_{extract_M}, T) 
\end{aligned}
\end{equation}    
}

Where $I_{hist}$ and $I_{med}$ are the elements from the patient's medical history and medication list most relevant to the patient's message. $P_{extract_H}$ and $P_{extract_M}$ are topic-specific information extraction prompts for history and medications. EHR reasoning agents then generate EHR-specific followup questions $\hat{Q}_{EHR} = \hat{Q}_{hist} \cup \hat{Q}_{med}$ as follows. Here, $P_{hist}$ and $P_{med}$ guide $f$ on generating questions related to $I_{hist}$ and $I_{med}$, respectively. 

{
\small

\begin{equation}
    \begin{aligned}
        &\hat{Q}_{hist} = f(T, I_{hist}, P_{hist}, k) \\
        &\hat{Q}_{med} = f(T, I_{med}, P_{med}, k)
    \end{aligned}
\end{equation}
}

\noindent \textbf{Differential Diagnostic Agents: } 
Providers who read and respond to patient messages often mentally perform a differential diagnosis before coming up with the follow-up questions \cite{ferri2010ferri}. Specifically, providers will (i) decide what could be wrong with the patient and (ii) ask questions to rule out various diagnosis hypotheses. Inspired by this mental framework, FollowupQ uses differential diagnostic agents to generate a set of possible patient diagnoses $D_{diff}$, then generate follow-up questions based on these potential diagnoses, i.e.follow-up questions to rule out each diagnosis $d_i \in D_{diff}$.

Differential diagnostic agents first compute a pseudo-differential diagnosis by identifying the $k$ best and worst-case diagnoses for a given patient, using prompts $P_{best}$ and $P_{worst}$ respectively.

\begin{equation}
\begin{aligned}
    &D_{diff} = f(T, P_{best},k) \cup f(T, P_{worst}, k)
\end{aligned}
\end{equation}

Thus, $D_{diff}$ is the union of the potential diagnoses produced by $f$ under the assumptions of the best and worst case scenarios. This strategy is motivated by the following observation we made in our preliminary work:
providers cast a wide net for gathering relevant information. Next, differential diagnostic agents iteratively build the question set $\hat{Q}_{{D_{diff}}} = \hat{Q}_{d_1} \dots \cup \dots \hat{Q}_{d_{|D_{diff}|}}$, which has targeted questions to rule out each diagnosis $d_i \in D_{diff}$. 
For a given possible diagnosis $d_i$ we compute $\hat{Q}_{d_i}$ as follows. 

\begin{equation}
    \hat{Q}_{{d}_i} = f(T, d_i, P_{rule-out}, k)
\end{equation}

Where $f$ outputs a question set of size $k$, using the prompt $P_{rule-out}$ to guide questions to see if the patient is suffering from potential diagnosis $d_i$. By taking the union of question sets about each potential diagnosis, differential diagnostic agents produce the set of questions $\hat{Q}_{{D_{diff}}}$.

\noindent \textbf{Message Clarification Agents: } FollowupQ's third type of agent is a set of message clarification agents to increase the clarity of different aspects of the patient's message. \textbf{Symptom inquiry} agents extract symptoms from the message and ask clarifying questions about each symptom as needed (e.g., location of abdominal pain). \textbf{Self-treatment} agents ask patients to elaborate on how they are treating their symptoms (e.g. over-the-counter medications to treat pain). \textbf{Temporal reasoning} agents generate questions to increase clarity in the timeline of presented symptoms, (e.g., duration and frequency of pain). \textbf{Message ambiguity} agents target reducing overall ambiguity of the message (e.g. ``tell me more about what you mean by you are feeling off."). For each clarification agent $clar_i$,

\begin{equation}
    \hat{Q}_{clar_i} = f(T, P_{clar_i}, k)
\end{equation}

Where $P_{clar_i}$ is a prompt specific to clarification agent $clar_i$. Taking the union over all clarification agents, we end up with question set $\hat{Q}_{clar}$. 

Our final question pool $\hat{Q}_p$ consists of questions generated by the differential diagnostic, medical chart, and clarification agents.  

\begin{equation}
\begin{aligned}
\hat{Q}_p = \{\hat{Q}_{D_{diff}}, \hat{Q}_{EHR}, \hat{Q}_{clar}\}
\end{aligned}
\end{equation}

\subsection{Question Filtration}\label{sec:filter}
The generation step produces $|\hat{Q_p}|$ questions, which based on hyperparameter choices in the FollowupQ framework (e.g. $k$ ) may produce a $\hat{Q_p}$ too large for a patient to answer. So, as ablation to our core experiments, we investigate ways to systematically reduce the size of $\hat{Q_p}$ to a target size $k$. Specifically, our framework performs question de-duplication and top-k question selection. Our question de-duplication framework uses LLMs to filter non-unique inquiries from $|\hat{Q_p}|$. This is a crucial step as different agents may request the same information in the context of different thought processes. Our top-k selection agent takes the de-duplicated question list and selects the k most important questions to ask the patient. Details of the question filtration method are presented in Appendix \ref{apn:filtration_step}.

%% file: sections/5_ExperimentalSetup.tex
\section{Experimental Setup}

\subsection{Dataset Details}

We evaluate our methods on a novel benchmark \textbf{FollowupBench} (FB) which contains two asynchronous Portal Message datasets: \textbf{FB-Real} and \textbf{FB-Synth} which are described in Table \ref{tab:dataset}. 

\textbf{FB-Real} consists of real messages and EHR records sent from adult patients to their providers between January 2020 and June 2024 at a large university medical center in the United States. From a corpus of over 500k messages, we perform a multi-step filtering process, including extensive human evaluation, to select messages which ensure each patient in our dataset is symptomatic and received a provider response containing follow-up questions. Human reviewers ensure that all clarification questions included in the ground truth are in scope for an AI system to generate (i.e. are grounded to the message or chart and not to prior in-person patient-provider interaction). Additionally, human reviewers ensure all questions are specific to symptom clarifications (i.e. we do not aim to generate logistical questions related to scheduling, insurance coverage, or medication refills). All extracted ground truth questions are broken down into single-topic questions to promote granular evaluation of NLP systems (e.g. \textit{"Do you have any fever or cough?"} is converted to the following two questions: "1. Do you have any fever? 2. Do you have any cough?"). As our human review process is expensive and extremely time-consuming, we limit FB-Real to 150 unique patient messages and the corresponding EHR data of those 150 patients. We provide extensive details of our data curation process in Appendix \ref{apn:fb_real}. As FB-Real contains protected health information, it can not be shared publicly.

\textbf{FB-Synth } is a semi-synthetic dataset consisting of 250 (medical chart, patient message) pairs with over 2,300 follow-up questions written by a team of 9 physicians, nurses, and medical residents at a large university medical center. This patient data was created by first sampling a random, de-identified medical chart from the corpus of patients used to create FB-Real. Then, we create a synthetic message using a grounded message generation technique inspired by \citet{gatto2024incontextlearningpreservingpatient}. Interestingly, FB-Synth has a higher mean number of questions per sample compared to FB-Real (9.3 vs 3.4). This is likely due to various factors including (a) research subjects often act differently under observation, known commonly as the Hawthorne Effect \cite{mccambridge2014systematic} (b) providers may request more information when they have more time to reflect on a given patient's needs. The latter point is suggestive of FB-Synth potentially being closer to the target set of questions an AI system should strive to generate.  
We make this dataset available to the research community. Please see Appendix \ref{apn:fb_real} for additional FB-Synth details.

\subfile{../tables/dataset_details}

\subsection{Baseline Methods \& Models}

We compare FollowupQ to the following baseline prompting approaches: 

\noindent \textbf{Unbounded-Generation:} We provide an LLM with both patient message and EHR data, and prompt it to write as many questions as necessary to clarify ambiguity and seek missing details. We explore unbounded generation in the context of both 0-shot and few-shot prompting. 

\noindent \textbf{$k$-Question-Generation:} The same prompt as unbounded-generation, but with specific guidance to output $k$ questions. This allows us to explore performance correlation with the scale of the generated set. For large values of $k$, this experiment explores the intrinsic limits of LLMs to reach long-tail questions. We explore $k=40$ in this study as the mean number of questions output by our FollowupQ framework before filtration is $\approx 35$. 

\noindent \textbf{Long-Thought Generation: } Given the recent success of long-thought models on complex reasoning tasks \cite{deepseekai2025deepseekr1incentivizingreasoningcapability} we explore the performance of models that generate follow-up questions after significant Chain-of-Thought output. This baseline is explicitly instructed to generate as many questions as it sees fit.

Due to the sensitive nature of FB-Real, we run our experiments on a secure computing cluster with no access to the internet and a single NVIDIA A40 GPU. We thus perform all experiments using 4-bit quantized versions of the following models due to computational restraints: (i) Llama3-8b \cite{grattafiori2024llama3herdmodels}, (ii) Llama3-8b Aloe, a Llama3 variant trained on healthcare data, \cite{gururajan2024aloe}, (iii) Qwen2.5-32b-instruct \cite{qwen2025qwen25technicalreport} --- the largest model we have available on our computing platform, and (iv) Qwen2.5-32b distilation of DeepSeek R1, for long-thought baseline. Appendix \ref{apn:hyperparameters} contains additional modeling details and Appendix \ref{apn:prompts} contains the relevant prompts.

\subsection{LLM As Judge}
\label{sec:llm-as-judge}

In Section \ref{sec:formal_metric} we formally define the metrics employed in our evaluation. Each metric depends on determining if a provider and LLM-generated question are requesting the same information. To detect matching questions, we employ a fine-tuned PHI-4-14b \cite{abdin2024phi4technicalreport} based LLM-as-Judge \cite{llm_as_judge} framework to do a pairwise comparison between all true and generated follow-up questions. We use a test set (n=100 question pairs) hand-labeled by a family medicine physician with over twenty years of experience.  We find that our judge model can detect matching question pairs that elicit the same information with a macro F1-score of 0.87. Appendix \ref{apn:judge} presents additional details of our Judge model, including fine-tuning procedure, example matches, prompts, and additional LLM performance on our test set.

%% file: tables/dataset_details.tex
\begin{table}[]
\centering
\resizebox{\columnwidth}{!}{%
\begin{tabular}{@{}cccc@{}}
\toprule
\textbf{Dataset}   & \textbf{\# of Messages} & \textbf{Total/Mean \# of Questions}  & \textbf{Mean \# of Sentences } \\ \midrule
\textbf{FB-Real}  & 150                     & 514 / 3.4                                           & 5.3                                       \\
\textbf{FB-Synth} & 250                     & 2,336 / 9.3                                          & 6.5                                       \\ \bottomrule
\end{tabular}%
}\caption{Dataset statistics for FollowupBench (FB). FB-Real has fewer questions on average as they were composed in a live, time-sensitive work environment.} 
\label{tab:dataset}
\end{table}

%% file: sections/6_Results.tex
\subfile{../tables/table_A}

\noindent \textbf{FollowupQ Significantly Outperforms Baseline Methods on Followup Question Generation: } In Table \ref{tab:main_results_real} we show that FollowupQ (Llama3-8b) achieves a mean RIM score of 0.62 on FB-Real while generating 36 questions. This is a 22-point increase in RIM compared to comparable zero and few-shot baselines using Llama3-8b --- showing the effectiveness of FollowupQ. Although Llama3-8b-Aloe achieves an even higher score of 0.64,  it also generates more additional questions. However, we note that Llama3-8b-Aloe uniquely struggled to follow instructions that pertain to output set size. This can result from it's training data \cite{luo2025empiricalstudycatastrophicforgetting}. We thus consider FollowupQ (Llama3-8b) as our top-performing model on FB-Real dataset. 

Interestingly, we find that while baseline solutions do see an increase in performance when encouraged to generate more questions (i.e. Unbounded $\rightarrow$ 40-question generation), they still struggle to match performance of FollowupQ. In other words, while FB-Real averages 3.4 follow-up questions per sample, our results demonstrate that even encouraging the LLM to generate over 10x the number of questions written by a real doctor does not solve this problem --- motivating a more intricate solution. In summary, FollowupQ's ability to generate diverse questions provides significant improvements over baseline LLMs in both zero and few-shot settings.

On the FB-Synth dataset, we find that FollowupQ with Llama3-8b provides a 5-point improvement over the closest baseline. When employing Qwen-32b, we find that the increase in performance is more subtle, but that FollowupQ still outperforms zero and few-shot baselines while generating fewer questions on average.

\begin{figure}[!t]
    \centering
    \includegraphics[width=\linewidth]{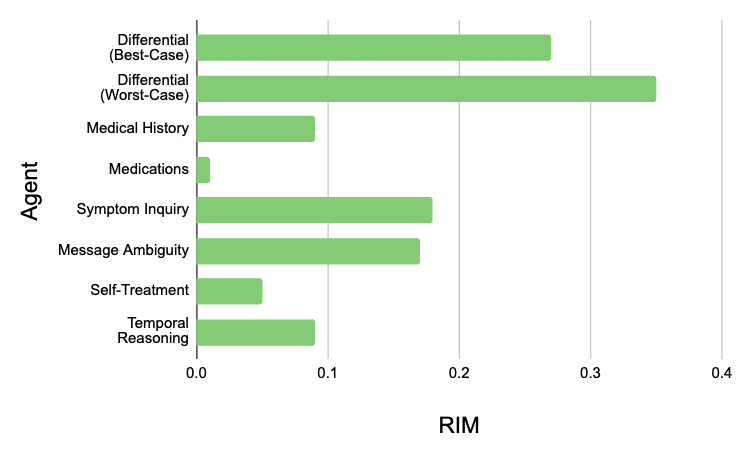}
    \caption{Per-Agent Performance on FB-Real from FollowupQ (Llama3-8b). We find that most of the performance comes from agents trying to rule-out the worst case scenario for a patient.} 
    \label{fig:per_agent_performance}
\end{figure}

\noindent \textbf{FollowupQ Reduces The Number of Information-seeking Messages providers Need to Send by 34\%: } In Table \ref{tab:full_match} we compare each model's ability to achieve RIM score of 1.0 on FB-Real --- indicating that all provider follow-up questions were captured by the model. We find that FollowupQ provides a significant 19\% increase over the nearest baseline. This suggests that when patients respond to all FollowupQ questions, providers need to request additional information in 34\% fewer symptomatic inquiries, effectively reducing their workload. FollowupQ’s provider-centric question-generation strategy produces a broader range of question types, improving performance over baseline LLMs, which often overlook less common concerns in a provider’s differential diagnosis.

\noindent \textbf{FollowupQ Surfaces Patterns in providers' Thought Process: }
In Figure \ref{fig:per_agent_performance} we show the RIM score of each specialist agent in FollowupQ on FB-Real. We find that most performance comes from the \textit{Differential (worst-case)} agent, displaying how follow-up questions are often concerned with ruling out worst-case scenarios as they would require urgent care. However, a non-trivial amount of inquiries come from other specialist agents focusing on medications, timeline clarification, and ambiguity clarification. Notably $\approx 10\%$ of performance came from inquiries concerning the patient's EHR, highlighting the importance of EHR data in personalized follow-up question generation. Thus, Figure \ref{fig:per_agent_performance} not only demonstrates the interpretability of our framework, but also surfaces underlying thought processes of the providers.

\subfile{../tables/fully_matched}

\noindent \textbf{Performance After Filtration Step}
We explore filtration on FB-Real using the best performing LLM, Llama-3-8b. In this experiment, we filter $|\hat{Q_p}|$ to a size of 10. The choice of k=10 is motivated by both conversations with clinical experts and the mean number of questions asked in our synthetic dataset (see Table \ref{tab:dataset}).  We find that our top-performing model, on average, generates 36 questions. After de-duplication, we go from $36 \rightarrow 22$ questions per sample with a mean RIM score of $0.62 \rightarrow 0.57$. 

Once we perform Top-10 filtering on the de-duplicated set, the RIM score drops from $0.57 \rightarrow 0.42$. This is likely not due to the quality of the questions, but rather our inability to model the subjective preferences of the provider who wrote the ground truth responses, as dialogue data is highly prone to response variability. However, our result still demonstrates the highest 10-question performance across all experiments. Future works may aim to model provider-specific agents to personalize the selection process.

%% file: tables/table_A.tex
\begin{table}[]
\centering
\resizebox{\columnwidth}{!}{%
\begin{tabular}{@{}llll@{}}
\toprule
             & Llama3-8b        & Llama3-8b-Aloe   & Qwen-32b         \\ \midrule
Zero-Shot-U  & 0.35 / 11 & 0.35 / 17 & 0.35 / 10 \\
Few-Shot-U   & 0.29 / 9  & 0.33 / 19 & 0.36 / 10 \\
Zero-Shot-40 & 0.40 / 30 & 0.40 / 30 & 0.41 / 35 \\
Few-Shot-40  & 0.37 / 35 & 0.36 / 35 & 0.45 / 37 \\
Long-Thought & -         & -         & 0.31 / 10 \\
Follow-up Q  & \textbf{0.62 / 36} & 0.64 / 58 & 0.54 / 35 \\ \bottomrule
\end{tabular}%
}\caption{The (Mean RIM / Mean Number of Questions Asked) for each experiment on FB-Real. We find that Llama3-8b achieves the best performance when balancing number of questions generated vs RIM. }
\label{tab:main_results_real}
\end{table}

\begin{table}[]
\centering
\resizebox{\columnwidth}{!}{%
\begin{tabular}{@{}clll@{}}
\toprule
             & Llama3-8b        & Llama3-8b-Aloe   & Qwen-32b           \\ \midrule
Zero-Shot-U  & 0.30 /  11 & 0.30 / 17 & 0.30 / 11   \\
Few-Shot-U   & 0.24 /  8  & 0.27 / 18 & 0.27 / 10   \\
Zero-Shot-40 & 0.34 /  30 & 0.35 / 31 & 0.43 / 36   \\
Few-Shot-40  & 0.34 /  35 & 0.35 / 35 & 0.41 / 38   \\
Long-Thought & -          & -         & 0.27 / 11   \\
Followup-Q   & \textbf{0.48 /  35} & 0.51 / 59 & 0.44 / 34 \\ \bottomrule
\end{tabular}%
}\caption{The (Mean RIM / Mean Number of Questions Asked) for each experiment on FB-Synth. We find FB-Synth to be slightly more challenging than FB-Real but with Llama3-8b still achieving superior performance.  }
\label{tab:main_results_synth}
\end{table}

%% file: tables/fully_matched.tex
\begin{table}[!t]
\centering
\resizebox{0.7\columnwidth}{!}{%
\begin{tabular}{@{}cc@{}}
\toprule
\textbf{Method}       & \textbf{Message Reduction \%} \\ \midrule
\textbf{Zero-Shot-40}       &         0.15                       \\
\textbf{Few-Shot-40}        &         0.13                   \\
\textbf{Followup-Q}         &         0.34                       \\ \bottomrule
\end{tabular}%
}\caption{Number of samples fully matched by each method from our Llama3-8b experiments. This table thus demonstrates potential for workload reduction, with Followup-Q having the potential to reduce need for sending symptom clarification messages by up to 34\%. }
\label{tab:full_match}
\end{table}

%% file: sections/7_Discussion.tex
\section{Discussion}
In this study, we introduce a novel task and dataset for advancing information-seeking LLM agents in the context of asynchronous medical conversations in primary care. We find that off-the-shelf LLMs struggle to generate question sets written by real providers. This result motivates our FollowupQ framework, which takes a multi-agent approach to question generation targeting a diverse and domain-specific line of medical inquiry. Our results demonstrate that FollowupQ significantly outperforms baseline LLMs and has the potential to reduce the asynchronous messaging workload of healthcare providers.

\section{Limitations}

This work is limited in that we are only able to explore a limited number of LLMs due to the computational restrictions of our secure computing environment. Additionally, future works will explore the search for optimal FollowupQ hyperparameters leveraging insights from our preliminary results. This was considered out of scope for the current submission. 

Both a strength and weakness of our dataset is that it is sourced from a single hospital in a rural community. This is a strength as rural populations may be underrepresented in medical NLP datasets. However, this is a weakness as our patient population may be biased towards certain sub-populations and our providers may be biased towards asking questions common to patients they frequently care for at this single hospital. Another limitation of our dataset is that it may be the case that different doctors will respond to the same patient message with different sets of questions. This phenomenon is the product of a variety of factors including (i) where and when was the doctor trained? (ii) what past experiences have the doctor had? (iii) how busy is the doctor at the time of reading the message? However, we note this is an unavoidable characteristic common to any medical dialogue dataset. 

Finally, we highlight that we do not explore our methods in the context of synchronous medical dialogue. However, future works may be inspired by the FollowupQ framework to generate questions for synchronous dialogue. 

\section{Ethical Considerations}
This paper was conducted under IRB approval from the submitting authors' institution. We will add IRB information to the ethical considerations upon acceptance. The FB-Real dataset contains sensitive patient information that cannot be publicly shared. This dataset was only utilized in a secure computing environment and handled by researchers who have completed HIPAA training. 

While there are no risks to any human subjects as a result of this study, we highlight that real-world utilization of our framework may have potential risks to patients if certain precautions are not taken such as employing rail-guarded LLMs which prevent the generation of harmful or offensive content. Additionally, any medical NLP system utilizing patient data must consider patient data privacy policies and protect user data from being stored or utilized inappropriately.

%% file: sections/Appendix.tex
\newpage 
\section{Additional Results}

Below we show global question matching scores which describe the overall percentage of questions matched across all samples. Note that this is distinct from RIM which measures sample-level coverage. 
\begin{table}[!h]
\centering
\resizebox{\columnwidth}{!}{%
\begin{tabular}{llll}
\hline
             & Llama3-8b & Llama3-8b-Aloe & Qwen-32b  \\ \hline
Zero-Shot-U  & 0.34 / 11 & 0.34 / 17      & 0.38 / 10 \\
Few-Shot-U   & 0.28 / 9  & 0.33 / 19      & 0.35 / 10 \\
Zero-Shot-40 & 0.42 / 30 & 0.40 / 30      & 0.44 / 35 \\
Few-Shot-40  & 0.37 / 35 & 0.38 / 35      & 0.45 / 37 \\
Long-Thought & -         & -              & 0.32 / 10 \\
Follow-up Q  & 0.58 / 36 & 0.61 / 58      & 0.52 / 35 \\ \hline
\end{tabular}%
}\caption{FB-Real: (Percentage of total questions matched by each model / mean number of questions asked). }
\end{table}

\begin{table}[!h]
\centering
\resizebox{\columnwidth}{!}{%
\begin{tabular}{llll}
\hline
             & Llama3-8b & Llama3-8b-Aloe & Qwen-32b    \\ \hline
Zero-Shot-U  & 0.28 / 11 & 0.29 / 17      & 0.27 / 11   \\
Few-Shot-U   & 0.22 / 8  & 0.26 / 18      & 0.25 / 10   \\
Zero-Shot-40 & 0.32 / 30 & 0.34 / 31      & 0.41 / 36   \\
Few-Shot-40  & 0.32 / 35 & 0.33 / 35      & 0.40 / 38   \\
Long-Thought & -         & -              & 0.25 / 11   \\
Followup-Q   & 0.46 / 35  & 0.49 / 59      & 0.41 / 34 \\ \hline
\end{tabular}%
}
\caption{FB-Synth: (Percentage of total questions matched by each model / mean number of questions asked) }
\end{table}

\section{Modeling Details } \label{apn:hyperparameters}

\subsection{Hyperparameters}
For all of the experiments in Tables \ref{tab:main_results_real} and \ref{tab:main_results_synth}, we use temperature = 0.6 with the remaining parameters left as default. 

FollowupQ has various values of $k$ used for each agent. We choose a reasonable $k$ value for our experiments based on our conversations with providers but plan to explore a broader hyperparameter search in future works. EHR Agents: $k=1$ (i.e. 1 question for medical history, 1 question for medications). Differential Diagnostics: $k=3$ per diagnosis. Message Clarification Agents $\rightarrow$ Symptom inquiry agents write $k=2$ questions per symptom. Message ambiguity and temporal reasoning use $k=3$. Self-treatment agents use $k=2$. We use a temperature of 0.6 during inference. 

\subsection{Experimental Cost}
We use an NVIDIA A40 GPU to run our experiments. We estimate our results required between 100-150 hours of computing time. 

\section{Filtration Step Details} \label{apn:filtration_step}

\noindent \textbf{De-Duplication:} For a given set of follow-up questions $\hat{Q}$, we first embed each $\hat{q_i} \in \hat{Q}$ using SBERT \cite{reimers-gurevych-2019-sentence} and compute $k$ question clusters $ X = \{x_1, \dots, x_k\}$ using K-Means clustering. We then prompt a de-duplication agent to reduce each cluster into a set of atomic questions about a single topic $\hat{Q}_{x_i} = f(x_i, P_{redundant})$.

The filtered question list $\hat{Q}_{filter}$ is the union of filtered question clusters $\hat{Q}_{x_i} \forall x_i \in X$. Note we de-duplicate on a per-cluster basis to improve accuracy when employing smaller, open-source LLMs, which we find are unable to properly filter large question lists in a single shot. By clustering, we reduce the complexity of the task by trying to maximize the overlap of questions in a given filtering step, making it clear what elements are redundant. 

\noindent \textbf{Top-K Selection: } Finally, we design a top-k filtration agent to return the top-k most important questions to reducing ambiguity in the patient's message. 

\begin{equation}
\hat{Q} = f(T, \hat{Q}_{filter}, P_{top_k})    
\end{equation}

This agent takes the set of redundancy-filtered questions $\hat{Q}_{filter}$ and uses the top-k filtering prompt $P_{top_k}$ to generate the final set of predicted questions $\hat{Q}$.

Note that for both De-Duplication and Top-K Sampling, we use Qwen2.5-32b-instruct to perform these tasks.

\section{Auto-Evaluation Details} \label{apn:judge}

Results in this study use our metric, Requested Information Match (RIM), to compare model performance. RIM requires a pairwise comparison between all true and predicted question sets. Recall that a pair of matching questions means that the LLM's response should elicit the same information as the provider's question. In this section, we describe (i) a Physician-crafted test set (ii) prompts (iii) data generation \& fine-tuning procedure for auto-evaluation. 

\subsection{Physician-Crafted Test Set}
Our auto-eval test set was created by sampling 50 random questions from FB-Synth. The authors then sourced challenging candidates for positive and negative question matches for each question. By challenging we mean (i) non-matching candidates with high token overlap and (ii) matching candidates with high semantic similarity but potential structural differences. This design choice helps thoroughly vet the model, as questions that are near exact duplicates or on completely different topics are easy for a model to detect. 

Finally, a family physician with 20+ years of experience annotated our test set and the resulting match data included 56 matches and 44 non-matches. We provide this dataset as a supplement to this work. Please consider a few samples from our Auto-Eval test set. 

\noindent \textbf{Provider:} Was your workout more intense than usual?  

\noindent \textbf{LLM :}Have you been exercising? 

\noindent \textbf{Match:} No
\\

\noindent \textbf{Provider:} Does it hurt to touch? 

\noindent \textbf{LLM :} If you apply pressure on it with your fingers does the pain increase?

\noindent \textbf{Match:} Yes

\subsection{Prompt} \label{apn:judge_prompt}

\begin{promptbox}
    
        You are presented with two medical questions written by two different doctors. Determine if the answer to question B will elicit the information required to answer question A. Keep in mind that question B should be equally as specific or vague as question A in order for two questions to elicit the same information. 
        
        Output yes or no and nothing else. 
        
        Question A: \{A\}
        
        Question B: \{B\}
        
        Answer:
        
\end{promptbox}

\subsection{Data Generation \& Fine-Tuning Procedure}
Our evaluation procedure can be computationally expensive due to the need for pair-wise comparison between predicted and ground truth question sets. Thus, we employ PHI-4 \cite{abdin2024phi4technicalreport} as our evaluator as it is a smaller LLM (only 14B parameters) and is separate from the models used in our main results table thus avoiding any biases. 

To help PHI-4 better understand our task, we perform synthetic data generation to fine-tune the model using the prompt in Appendix \ref{apn:judge_prompt}. Specifically, we write five canonical examples of question matching in the form of contrastive learning examples and teach an LLM to generate similar data points based on randomly sampled medical symptoms. Consider one canonical example. 

\begin{verbatim}

Root: What did you stub your toe on?
Positive: To be clear, did you stub your
toe on something? What exactly 
did you hit?
Negative: Does your toe hurt?

## Task ##
-> Now, write a new contrastive sample,
this time about {topic}. 
\end{verbatim}

We use such examples to generate new contrastive data points about \{topic\}. We consider the root samples to be provider questions and the positive / negative samples to be LLM-generated questions for fine-tuning. We generate 1,000 contrastive samples, which leads to 2,000 fine-tuning examples for our judge model. Note that we perform n-gram de-duplication with our test set (n=5) to ensure we are not accidentally generating any test samples. The result of our fine-tuning can be found below.

\begin{table}[!h]
\centering
\resizebox{\columnwidth}{!}{%
\begin{tabular}{llll}
\textbf{Model}     & \textbf{Precision} & \textbf{Recall} & \textbf{F1} \\
\toprule 
Phi-4 (Base)       & 0.86               & 0.86            & 0.85        \\
Phi-4 (Fine-Tuned) & 0.88               & 0.87            & 0.87       
\end{tabular}%
}
\end{table}

We find that this gives us a 2-point boost on our physician-crafted test-set. We additionally note that empirical inspection of model outputs shows strong performance in identifying question matches. We fine-tune the model using Q-LoRA \cite{hu2021loralowrankadaptationlarge, dettmers2023qloraefficientfinetuningquantized} the SFT trainer from TRL \footnote{\url{https://huggingface.co/docs/trl/en/sft_trainer}} for 1 epoch with a learning rate of 2e-5 and batch size of 32. 

\section{FollowupBench Dataset Details} \label{apn:fb_real}
In this section, we outline (i) how we created FB-Real from asynchronous patient conversations sourced from an online EHR portal and (ii) how we created FB-Synth using semi-synthetic data and real provider follow-up questions. 

\subsection{Creation of FollowupBench-Real}

\begin{figure}[!h]
    \centering
    \includegraphics[width=\linewidth]{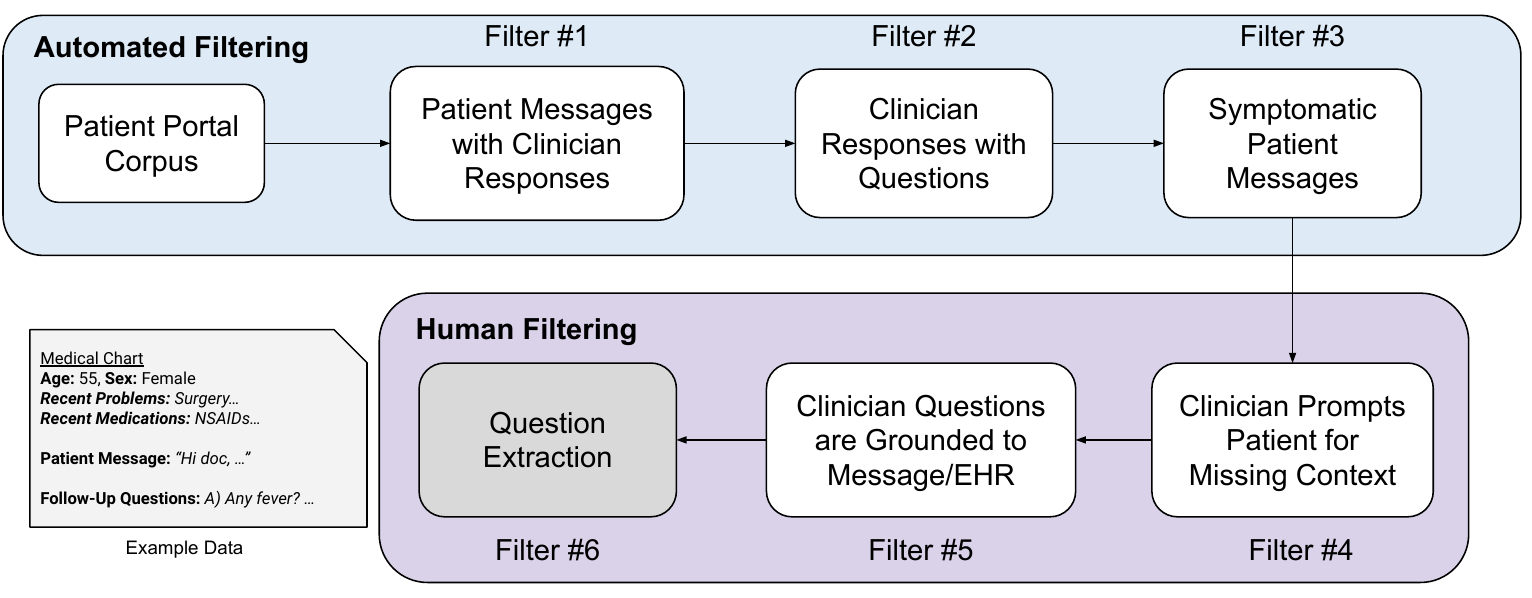}
    \caption{Visualization of the data collection process for FollowupBench-Real. }
    \label{fig:viz_fbreal}
\end{figure}

The goal of this study is to explore the capacity of LLMs to generate follow-up questions that pertain to any \textit{missing context} required to make an informed decision. Thus, we aim to focus on symptomatic patients who do not provide the full breadth of information required to be properly diagnosed. We visualize our sample selection pipeline in Figure \ref{fig:viz_fbreal}. First, we filter for patient-initiated conversations that received at least one provider response (Filter \#1). Next, we filter for conversations where the provider response contains at least one question mark `?' (Filter \#2). Finally, in an effort to emphasize symptomatic patients and de-emphasize logistical inquiries such as medication or appointment-centric messages, we employ a BERT-based \cite{devlin-etal-2019-bert} symptom extraction model introduced in \cite{raza2022large}. Based on this symptom tagger, we only consider messages that contain at least one symptom (Filter \#3). Based on these criteria, we have a pool of eligible symptomatic conversations. Our final filtering step (Filter \#4-6) leverages human annotation and is described in the following section.

\subsection*{Human Data Annotation}
From the pool of eligible conversations, two human annotators were tasked with selecting samples that met the following two criteria. 

\paragraph{Criteria \#1: } The provider response must prompt the patient for missing information about their symptoms. Any responses that solely pertain to logistical matters such as medication refills or appointment scheduling were not to be included. 

\paragraph{Criteria \#2: } The follow-up questions asked by the provider must be grounded to either the patient's message or their Chart/EHR. In other words, questions that appear based on information from a prior in-person encounter are deemed out-of-scope for the LLM and are thus not included in the final dataset for the sake of fairness. Additional out-of-scope message types are (i) those based on images submitted by patients (which we do not have access to) (ii) temporally dependent questions such as ``how are you feeling now?" which presumes significant time has elapsed since the message was processed (iii) questions which are healthcare suggestions (e.g. would you like to try this treatment?). 

If the human annotators deem a provider response meets both Criteria \#1 and Criteria \#2, we include the sample in our evaluation. Once a sample is considered eligible, the human annotator is tasked with extracting the questions from the provider response to be used later as our ground truth test set for LLM evaluation. Note that provider responses that contain both eligible and ineligible questions are still utilized, with all out-of-scope questions being excluded during evaluation. Our human annotators sampled messages until the target evaluation set size of (n=150) was reached. Note that for the sake of facilitating the auto-evaluation task, the human annotators were allowed to slightly rephrase the provider questions to be complete sentences. 

Our human annotators are senior Ph.D. students experienced in clinical NLP. Our data-inclusion criteria were developed in collaboration with a primary care provider with 20+ years of experience working in a large healthcare system. Our expert collaborator has significant experience in writing patient portal message responses. Each of our human annotators were trained over multiple sessions with our expert on how to identify quality samples for evaluation as well as parse provider messages into a list of questions. We highlight that, in general, medical expertise is not needed to determine if a question is predominantly logistical. For example, our human annotators were tasked with filtering out samples with questions such as ``we have an appointment available today, would you like to come in?" or ``which pharmacy did you want the medication sent to?". 

In summary, we use this process to curate the 150 samples used in FB-Real.

\subsection{Creation of FollowupBench-Synth}
Recall that FollowupBench-Synth is a semi-synthetic dataset containing (patient messages, linked EHR data, and follow-up question set) triplets. All patient messages in FB-Synth are 100\% synthetic. All EHR data is real and is randomly sampled from a de-identified pool of patient EHR records at a large university medical center. The follow-up questions were written by a team of providers who were presented with the synthetic messages and real EHR data. In the following sections, we describe each step in the synthetic data generation process.

\subsubsection{Synthesizing Patient Messages}

\begin{figure}
    \centering
    \includegraphics[width=\linewidth]{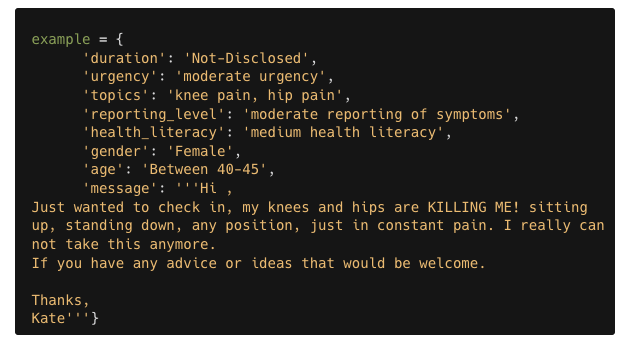}
    \caption{Example sample used in grounded generation. The goal is to map all non-message features to a novel synthetic message based on randomly sampled features. We provide few-shot examples in-context to guide the generation. }
    \label{fig:synth_ex}
\end{figure}

Inspired by \citet{gatto2024incontextlearningpreservingpatient}, our synthetic data generation pipeline uses grounded generation to promote sample realism. In this work, our goal is to map a randomly sampled set of categorical features into a new patient message. We use the following categorical features: (1) topics/symptoms (2) duration of symptoms (3) message urgency (4) reporting level, or how forthcoming the patient is with their symptoms (5) level of health literacy (6) age (7) gender. These categories were deemed relevant after discussion with clinical experts and helped promote sample diversity during generation. See Figure \ref{fig:synth_ex} for an example set of categories and corresponding messages.

\begin{figure}[!h]
    \centering
    \includegraphics[width=\linewidth]{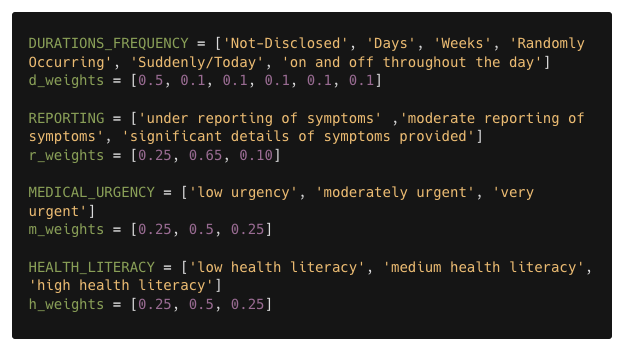}
    \caption{Synthetic data generation sampling categories for Duration, Reporting-Level, Urgency, and Health Literacy.     }
    \label{fig:categories}
\end{figure}

To generate a new sample, we (1) randomly sample 1-2 topics/symptoms from a list of common patient message topics curated by the authors and reviewed by a provider. The list of topics used for generation is provided as a supplement to this submission. For steps (2-5) we randomly sample from the values shown in Figure \ref{fig:categories} using the displayed categories and category weights. (6-7) Age and Gender are taken from the EHR data which will be matched with the synthetic message. 

We thus feed an LLM (Llama-3.3-70B-Instruct) 3 in-context samples of how to map a set of categorical variables to a realistic looking patient message. The three in-context samples were generated by the authors to mimic realistic patient message style. We use the outputs of this synthetic message generation framework in FB-Synth. 

\subsubsection{EHR Data Selection}

EHR data is randomly sampled from a pool of patients at a large university medical center. We take snapshots of patient EHRs corresponding to details outlined in Section \ref{sec:task-formal}. See Appendix \ref{apn:example_data} for an example of an EHR record used in this study. 

\subsubsection{Data annotation}
Pairs of synthetic patient messages and real EHR records were presented to practicing medical providers at a large university hospital for annotation. We discuss the details of the data annotation strategy below. 

\paragraph{Annotator Details} 
Nine providers were employed as annotators to create the FB-Synth dataset. Our annotators all work as nurses, medical residents, or primary care providers who have experience responding to patient portal messages. These annotators were recruited for our study via referral from a collaborating provider. All 9 annotators are female and provide healthcare services in the United States. All annotators were informed on how their data would be used as part of this study. 

\paragraph{Payment} Each provider was paid a fair wage of \$50 (USD) per hour to participate in this study. 

\paragraph{Annotation Details}
All providers employed in this study are familiar with writing follow-up questions to patient messages. Thus, annotators required little training. However, we highlight a few guidelines told to annotators during a brief live presentation: 

\begin{enumerate}
    \item Your job is to write a list of questions you wish the AI would have asked the patient before they click send! 
    \item Your goal is to ask questions that clarify a patient’s symptoms, self-treatment, and health status. 
    \item We are not looking for logistical questions! E.g. appointments, refills
    \item The list does not have to be perfect! There is no one right answer! Just use your best judgment. 
    \item You should write 3-10 questions per message. However, it is okay to go over or under this amount. 
\end{enumerate}

Annotators wrote follow-up questions at their own pace and reported their time spent accordingly. Most annotators in this study provided $\approx 3$ hours of annotation.

\section{Prompts} \label{apn:prompts}

In this section, we discuss the prompts used in our study. 

\subsection{Baseline Prompts}

\begin{figure}[!h]
    \centering
    \includegraphics[width=\linewidth]{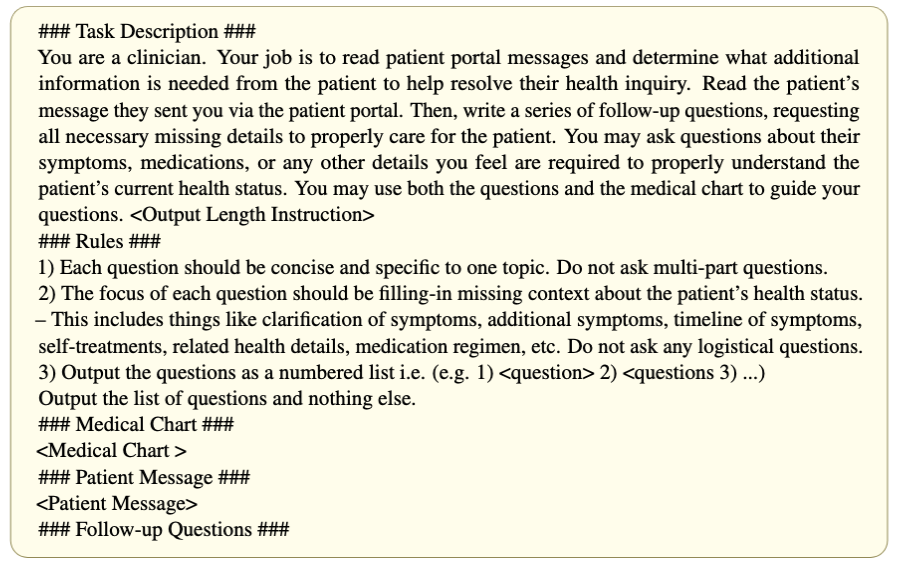}
    \caption{Core prompt used for zero-shot and few-shot baseline approaches.}
    \label{fig:baseline_prompt}
\end{figure}

Figure \ref{fig:baseline_prompt} shows the base prompt for all of our zero, few-shot, and long-thought experiments. Note that <Output Length Instruction> denotes the difference between unbounded and generation of K samples. The unbounded prompt states ``Write as many questions as you need" as where k-generation states ``write exactly k questions". The few-shot variants add examples in the prompt using the same set of instructions/rules.

\subsection{Multi-Agent Prompts}
FollowupQ uses 12 prompts at various stages of inference. Due to the considerable space this requires to display, we show the Differential Diagnostics-related questions as an example and refer the reader to the code base for the full suite of prompts. 

\begin{promptbox}
    \#\#\# Task Description \#\#\#
    
    You are a doctor. You have received a message from your patient. 
    
    Given the message, you suspect they have \#\#\# Suspected Issue \#\#\#
    However, you need more information before being sure. 
    Write \{k\} questions you would want to ask this patient to confirm if they do, or do not, have \#\#\# Suspected Issue \#\#\#. 

    \#\#\# Rules \#\#\#
    
    - Output the questions as a numbered list. E.g. 1) <question> 2) <question> ... 
    
    - Make questions as concise as possible. Do not ask multi-part questions. 
    
    - Only ask the {k} most important questions needed to clarify this specific diagnosis. 
    
    - Make the questions specific to ruling out your suspected issue. 

    \#\#\# Patient Message \#\#\#
    
    {msg}

    \#\#\# Suspected Issue \#\#\#
    
    \{suspected\_issue\}

    \#\#\# \{k\} Follow-up Questions \#\#\#:
\end{promptbox}
 
\section{Example Data}\label{apn:example_data}

Below is an example Chart, Message, and set of Follow-up questions. Note that this is a chart derived from MT-Samples \footnote{https://mtsamples.com/} which we used during early development of our annotation guidelines for FB-Synth (please note MT-Samples is not included in FB-Synth, and that this chart is for illustration only as we can not share real chart data here). The message is synthetic, and the follow-up questions are generated by a practicing triage nurse at our partner hospital, a large academic medical center.

\begin{promptbox2}
<Example Medical Chart>

\#\#\#Demographics\#\#\#

Age: 50

Gender: Male

\#\#\#Full Active Problem List\#\#\#

Diabetes mellitus type II - hypertension - coronary artery disease - atrial fibrillation - smokes 2 packs of cigarettes per day - Family history of coronary artery disease (father \& brother).

\#\#\#Recent Encounters (Max 10)\#\#\#

status post PTCA in 1995 by Dr. ABC 

\#\#\#Medications (Outpatient)\#\#\#

- Aspirin 81 milligrams QDay  

- Humulin N. insulin 50 units in a.m.  

- HCTZ 50 mg QDay.  

- Nitroglycerin 1/150 sublingually 

- PRN chest pain.
\end{promptbox2}

\begin{promptbox2}
<Patient Message>

Hey Dr. [Doctor's Last Name],
 
Getting a little nervous... I noticed I am breathing really rapidly and it's tough to catch my breath. On top of that, I'm coughing up blood, which has me pretty freaked out. This all started quite suddenly, and I'm not sure what's going on. Could we possibly arrange to see each other soon or talk about what I should do next? Can you call me?
\end{promptbox2}

\begin{promptbox2}
<Real provider Follow-up Questions> 

1) You say it started suddenly, but when exactly did it start? 

2) How much blood are you coughing up? A teaspoon? Flecs? 

3) Describe the color of the blood. (Dark vs light) 

4) Are you short of breath during any specific activities? E.g. sitting down vs moving. 

5) Are you able to talk in a full sentence without pausing? 

6) Any dizziness? 

7) Has this ever happened before? 

8) Have you noticed any blood in your stool? 

\end{promptbox2}